\documentclass[letterpaper, 10pt, conference]{template/ieeeconf}  % Comment this line out if you need a4paper
\usepackage[T1]{fontenc}
\usepackage{wrapfig}
\usepackage{booktabs}
\usepackage{amsmath}
\usepackage{amsmath} % assumes amsmath package installed
\usepackage{amssymb}  % assumes amsmath package installed
\usepackage[]{changes}
\definechangesauthor[name={Todor}, color=red]{tsv}
\definechangesauthor[name={Johannes}, color=cyan]{jsk}
\presetkeys%
    {todonotes}%
    {inline,backgroundcolor=yellow}{}
\usepackage{chngcntr}
\usepackage{siunitx}
\usepackage{tikz}
\usepackage{adjustbox}
\usepackage{url}
\usepackage{subfigure}
\usepackage{balance}

\IEEEoverridecommandlockouts                              % This command is only needed if 
\title{\LARGE \bf
Heterogeneous Full-body Control of a Mobile Manipulator with Behavior Trees}

\author{Marco Iannotta*, David C\'{a}ceres Dom\'{i}nguez*, Johannes A. Stork, Erik Schaffernicht, and Todor Stoyanov% <-this % stops a space
\thanks{Authors are with the Autonomous Mobile Manipulation Lab at the Center for Applied Autonomous Sensor Systems (AASS), Örebro University, Sweden {\tt [firstname.lastname]@oru.se}.
}%
\thanks{* denotes equal contribution. 
}%
}

\begin{document}

\maketitle
\thispagestyle{empty}
\pagestyle{empty}

%%%%%%%%%%%%%%%%%%%%%%%%%%%%%%%%%%%%%%%%%%%%%%%%%%%%%%%%%%%%%%%%%%%%%%%%%%%%%%%%
%%%%%%%%%%%%%%%%%%%%%%%%%%%%%%%%%%%%%%%%%%%%%%%%%%%%%%%%%%%%%%%%%%%%%%%%%%%%%%%%
\begin{abstract}
Integrating the heterogeneous controllers of a complex mechanical system, such as a mobile manipulator, within the same structure and in a modular way is still challenging.
In this work we extend our framework based on Behavior Trees for the control of a redundant mechanical system to the problem of commanding more complex systems that involve multiple low-level controllers.
This allows the integrated systems to achieve non-trivial goals that require coordination among the sub-systems.
\end{abstract}

%%%%%%%%%%%%%%%%%%%%%%%%%%%%%%%%%%%%%%%%%%%%%%%%%%%%%%%%%%%%%%%%%%%%%%%%%%%%%%%%
\section{\uppercase{Introduction}}
\label{sec:introduction}

Sophisticated mechanical systems, such as mobile manipulators, typically involve multiple sub-systems that have to be coordinated to create complex behaviors and achieve multiple goals at the same time.
Integrating the control of these sub-systems in a modular way within the same structure is still challenging, because multiple and heterogeneous controllers that run at different control frequencies are required.

In \cite{btsot}, we propose a novel method to combine a Stack-of-Tasks (SoT) control strategy \cite{5766760, 240390} with Behavior Trees (BTs) \cite{DBLP:journals/corr/abs-2005-05842}.
The former is an approach that allows a redundant (i.e., DOF > 6) robot to fulfill a number of prioritized goals.
The latter is a task switching structure functionally equivalent to Finite State Machines (FSMs) \cite{petter-bts, 6907656}, but that promises to address some of its limitations in terms of reactivity (i.e., ability to quickly and efficiently react to changes) and modularity (i.e., system’s components may be separated into building blocks, and recombined).  
In our approach, a BT and an SoT strategy run in parallel at different control frequencies.
On the basis of the current robot and environment state, the BT  periodically  configures  a  hierarchical  control  problem, which  is  then  solved  by  the  SoT  strategy  at  a  much  higher control frequency.
In \cite{btsot}, we demonstrate how this framework can be successfully applied to a 7-DOF manipulator to achieve complex goals in a modular, transparent and reactive way.

\begin{figure}[t!]
\centering
%never specify both height and width for a figure! Let Latex make sure you keep aspect ratio.
\includegraphics[width=0.97\linewidth]{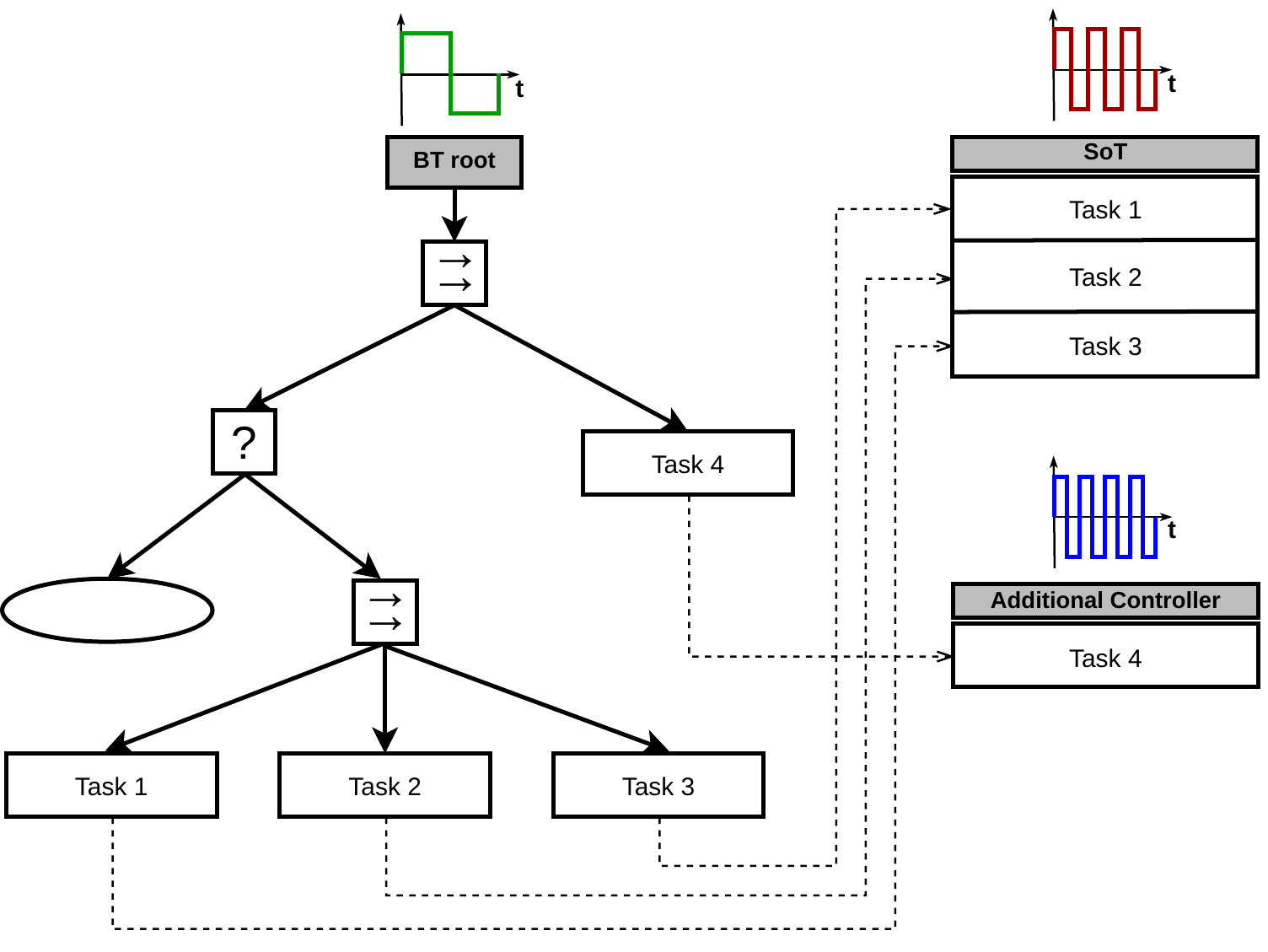}
\vspace{-0.2cm}% <--- this magic squeezes the space between figure and caption
\caption{Integration between the unified SoT-BT framework proposed in \cite{btsot} and additional controllers. The BT, the SoT approach and the additional controller run in parallel at different control frequencies.}
\label{fig:bt-multiple-controllers} %always label figures for reference within the text
\vspace{-0.2cm}% <--- this magic squeezes the space between the caption and the rest of the text
\end{figure}

In this paper, we extend the framework proposed in \cite{btsot} to the problem of commanding additional controllers for the handling of more complex 
systems.
Due to the nature of BTs, any behaviour can be encapsulated within a leaf node.
In this way, the use of standard BT leaf nodes allows to handle the tasks for different and heterogeneous controllers within the same BT (Fig. \ref{fig:bt-multiple-controllers}).
The proposed extension does not affect the advantages in terms of modularity, reactivity and transparency mentioned in the original work.
Moreover, handling heterogeneous controllers under the same policy allows the integrated systems to achieve non-trivial goals that would be difficult to attain by controlling the sub-systems in isolation.

We test our approach in simulation on a mobile manipulator, whose goal is to approach a table, and perform a pick and place operation, while the mobile platform is moving (Fig. \ref{fig:task-description-bt}). 

%%%%%%%%%%%%%%%%%%%%%%%%%%%%%%%%%%%%%%%%%%%%%%%%%%%%%%%%%%%%%%%%%%%%%%%%%%%%%%%%
\section{\uppercase{Background}}
\label{sec:related_work}

In this section we provide an overview of the used methodologies for the proposed approach. 
We start by describing the SoT control strategy in \ref{subsec:sot} and the BTs in \ref{subsec:bt}.
In Sec. \ref{subsec:sot-bt} we recall the main concepts of the framework we propose in \cite{btsot} to combine the SoT approach with BTs.

\subsection{Hierarchical Stack-of-Tasks Control Strategy}
\label{subsec:sot}

Each goal for a redundant mechanical system can be formulated in terms of minimizing a separate \textit{task} (or \textit{error}) function of the robot state, which can be regulated with an ordinary differential equation.
In case of multiple tasks, the corresponding equations can be sorted by priority and solved each in the solutions set of higher priority tasks (\textit{task-priority} or \textit{Stack-of-Tasks}).
Kanoun et al. \cite{5766760} proposed a prioritized task-regulation framework based on a sequence of quadratic programs (QP) that generalizes the previous \textit{task-priority} framework \cite{240390} to inequality tasks. In \cite{escande2014hierarchical} a more numerically efficient solution of the same problem was proposed.

The assumption behind this method is that it is always possible to reach a globally optimal solution by minimizing a quadratic error function.
In reality, this is not always the case and it often happens to deal with problems that require non-quadratic objectives.
In these situations, a single SoT might not be enough, because the control strategy might bring the robot to a disadvantageous configuration with respect to the global goal.
The only way to address these scenarios, without considering a more complex error function or a motion
planner, is to compose a sequence of local approximations (i.e., Stack-of-Tasks) that prevent the control strategy from falling into a local minimum.
In \cite{btsot} we propose to exploit Behavior Trees for this purpose. 

\subsection{Behaviour Trees}
\label{subsec:bt}

\begin{figure}[t!]
\centering

\subfigure[]{\includegraphics[width=0.5\linewidth]{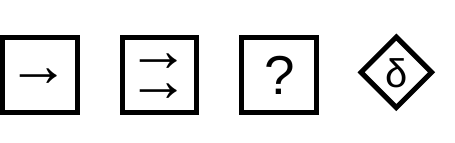}
\label{fig:control-nodes}
}

\subfigure[]{\includegraphics[width=0.5\linewidth]{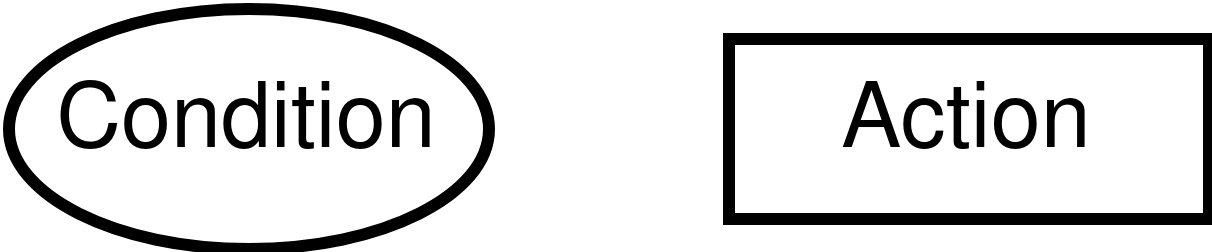}
\label{fig:execution-nodes}
}
\caption{(a) Control flow nodes, from left to right: Sequence, Parallel, Fallback and Decorator. (b) Execution nodes, from left to right: Condition and Action.}
\vspace{-0.5cm}
\end{figure}

Behaviour Trees are an alternative to FSMs that promise to address some of the limitations in modularity, re-usability and reactivity.  Following the terminology from \cite{DBLP:journals/corr/abs-1709-00084}, a BT is composed of a root node, and at least one internal node and one leaf node. The execution of a BT starts at the root node, which generates signals, or \textit{ticks}, with a given frequency, that are propagated to the children, down to the leaf nodes. A node is executed if and only if it receives a tick, and returns a status to the parent. The possible statuses are \textit{Running}, if the execution is under way, \textit{Success} if the node has achieved its goal, or \textit{Failure} otherwise.

Internal Nodes, or Control Nodes, can be of type \textit{Sequence}, \textit{Fallback}, \textit{Parallel},  or \textit{Decorator} (Fig. \ref{fig:control-nodes}). A Sequence Node propagates the tick to its children nodes from left to right and it returns \textit{Success} if and only if all its children return \text{Success}. A Fallback Node propagates the tick to its children nodes from left to right until one of them returns \textit{Success}. Note that, when a child returns \textit{Running}, the Sequence and the Fallback Nodes do not tick the next child if any. A Parallel Node propagates the tick to all its children simultaneously and it returns \textit{Success} if a set amount of them return \textit{Success}. Lastly, a Decorator node modifies the return status of a single child node with any custom policy.

Leaf Nodes, or Execution Nodes, include two categories (Fig. \ref{fig:execution-nodes}). When ticked, an Action Node performs a command. It returns \textit{Success} if the action defined by the command is correctly completed, \textit{Failure} if it has failed and \textit{Running} if it is ongoing. A Condition Node checks a proposition, returning \textit{Success} if it holds, \textit{Failure} otherwise.

%%%%%%%%%%%%%%%%%%%%%%%%%%%%%%%%%%%%%%%%%%%%%%%%%%%%%%%%%%%%%%%%%%%%%%%%%%%%%%%%
\begin{figure*}[t!]
\vspace{0.3cm}
\centering

\subfigure[]{\includegraphics[height = 0.38\linewidth]{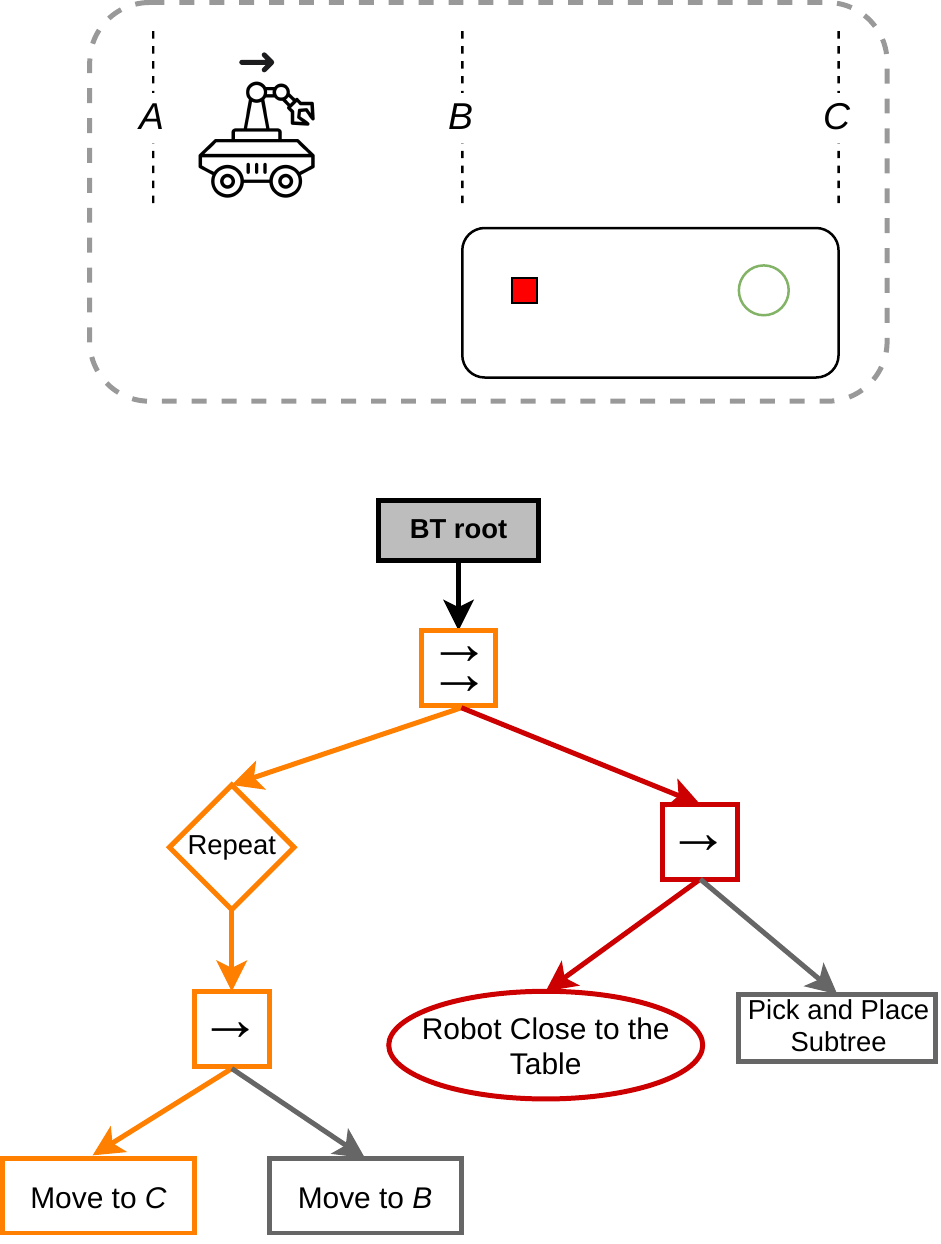}
\label{fig:task-description-bt-1}
}
\hspace{0.5cm}
\subfigure[]{\includegraphics[height = 0.38\linewidth]{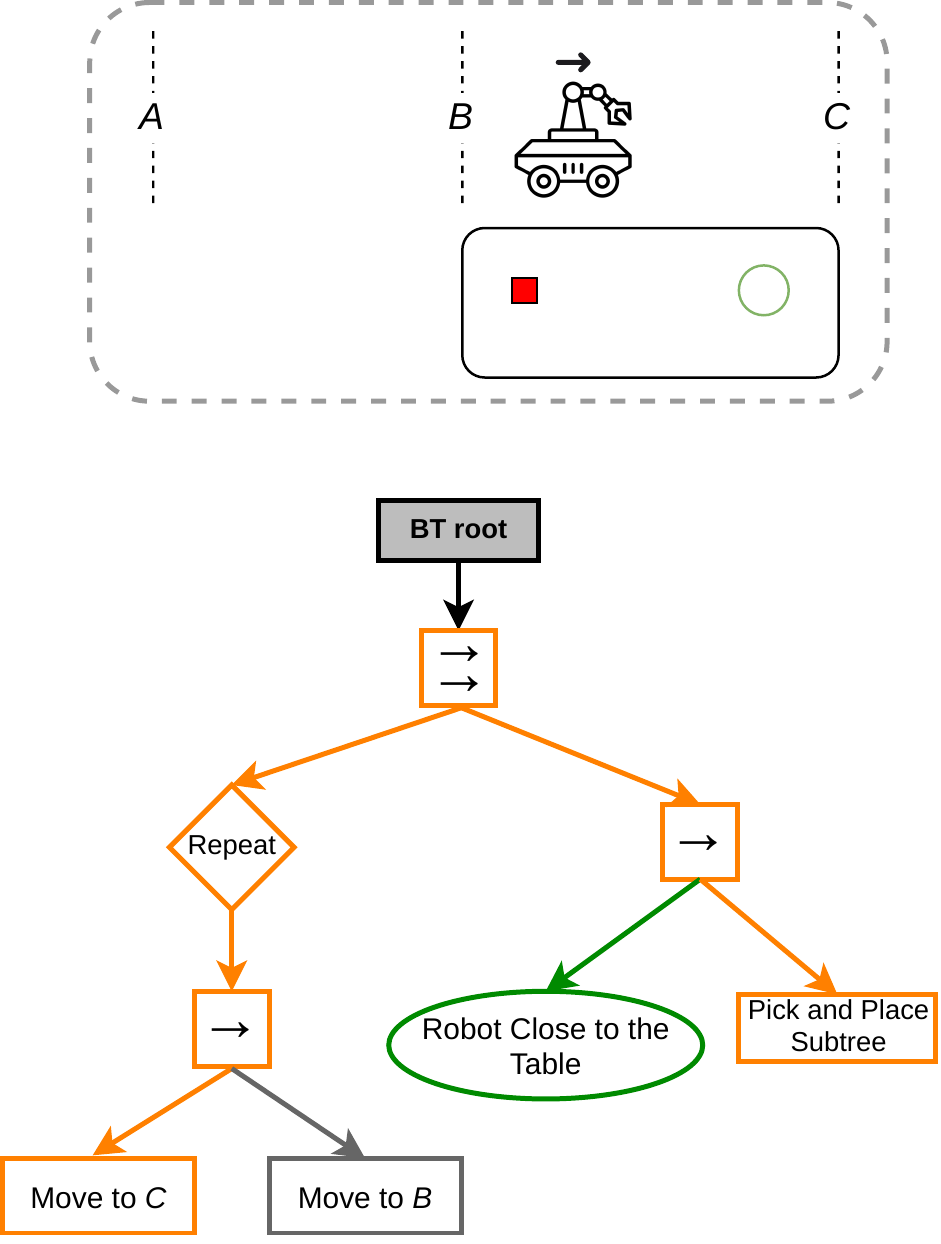}
\label{fig:task-description-bt-2}
}
\hspace{0.5cm}
\subfigure[]{\includegraphics[height = 0.38\linewidth]{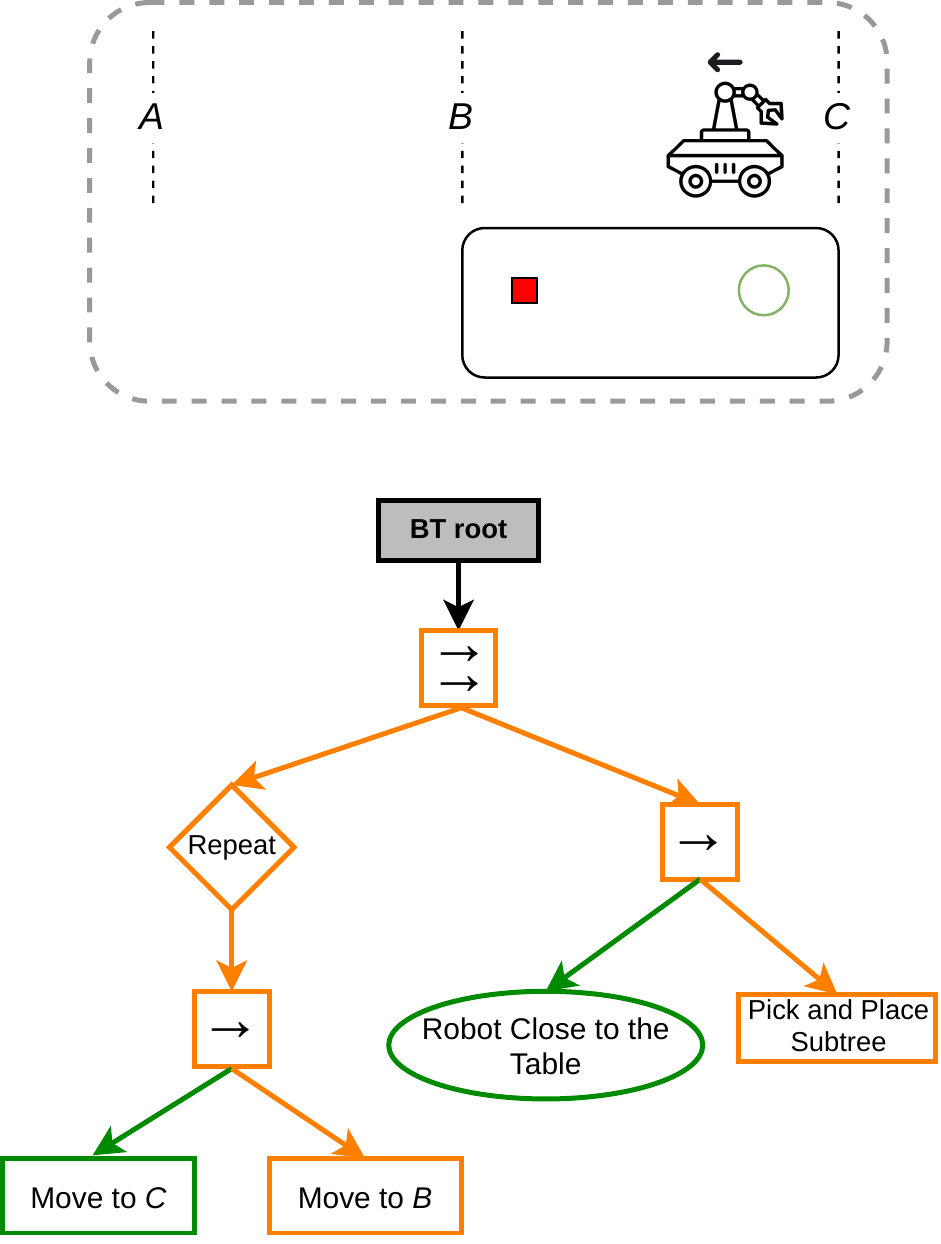}
\label{fig:task-description-bt-3}
}

\vspace{-0.25cm}
\caption{Task on which the approach is tested. The mobile manipulator has to approach a table, pick up a 30mm cube (in \textit{red}) and place it on the other side of the table (\textit{green circle}), while the mobile platform moves between the two extremes of the table (\textit{B} and \textit{C}). The bottom part of each figure shows the running BT for the scenario described in the correspondent upper figure. In the BTs, green denotes \textit{Success}, orange denotes \textit{Running}, red denotes \textit{Failure}, and grey denotes \textit{not ticked}.}
\label{fig:task-description-bt}
\vspace{-0.25cm}
\end{figure*}

\subsection{Combining SoT with BTs}
\label{subsec:sot-bt}

In \cite{btsot} we propose a framework to integrate one SoT control strategy with BTs, such that a BT is used to configure the hierarchical control problem to be solved by the SoT approach.
The aim is to exploit the flow of a BT to dynamically update the task functions in the SoT, while preserving the framework's transparency and modularity.
In this section, we briefly revise the main concepts and refer the reader to the original work for the details.

The basic idea is to have a BT and an SoT approach running in parallel at different control frequencies, and to use the BT for mapping a robot and environment state to a hierarchical control problem.
Once configured by the BT, this control problem is then solved by the SoT strategy at a much higher control frequency.
In this way, we can constantly update the task functions in the SoT by either setting new tasks, or removing the ones which are no longer necessary.
We define customized Action and Control Nodes in the BT to configure the hierarchical control problem. In each new Action Node, we set a single task (i.e. the associated \textit{task constraint} is added to the hierarchical control problem), instead of an entire Stack-of-Tasks, as it brings two main advantages.
On the one hand, it improves understandability, since the BT provides a clear way to visualize all the running tasks.
On the other hand, tasks that are in common for an entire subtree and not strictly related to the behavior defined by it (typically constraints, i.e. avoid collision with an object) can be defined only once, outside the scope of the subtree.
This makes each subtree more independent and easily re-usable in different contexts.

We use the new Control Nodes to remove the previously set tasks.
Task removal takes place immediately before the Control Node returns its final outcome (i.e., \textit{Success} or \textit{Failure}), without affecting the node behavior described in Section \ref{subsec:bt}.
The creation of new Control Nodes allows us to exploit these nodes for task removal when needed, without removing the possibility to use the standard ones just for controlling the flow of the BT.
%%%%%%%%%%%%%%%%%%%%%%%%%%%%%%%%%%%%%%%%%%%%%%%%%%%%%%%%%%%%%%%%%%%%%%%%%%%%%%%%

\section{\uppercase{Approach and evaluation}}
\label{sec:approach-evaluation}

In this paper we extend the framework presented in Sec. \ref{subsec:sot-bt} to the problem of commanding complex mechanical systems that are tracked by heterogeneous low-level controllers.

The key point of having the BT and the SoT approach running in parallel at different control frequencies is extended to the new controller.
The use of a BT as control policy allows us to easily perform the integration of the new controller, by exploiting the intrinsic modular nature of BTs when performing task composition.
As previously mentioned, in \cite{btsot} we define for convenience new Action and Control Nodes to respectively set and remove single manipulation tasks (e.g., follow a line or go to a position) in the SoT.
On the contrary, current works on BTs typically encapsulate entire operations that involve multiple goals in standard Action Nodes.
To show the ease of integration of our framework with every kind of controller and approach in the literature, we use standard Action Nodes to handle the tasks related to additional controllers.
However, the use of a control strategy similar to the SoT one, that decomposes complex goals in a set of tasks to be handled separately in the Action Nodes, would improve the overall modularity and transparency. 
The resulting framework is shown in Fig. \ref{fig:bt-multiple-controllers}.
At each tick, new tasks for the additional controllers are commanded in addition to the dynamic update of the hierarchical control problem for the SoT strategy.
In this way, we can coordinate the sequential or simultaneous execution of different controllers to achieve one or multiple goals at the same time.

\begin{figure*}[t!hb]
\vspace{0.3cm}
\centering

\subfigure[]{
\includegraphics[height = 0.2\linewidth]{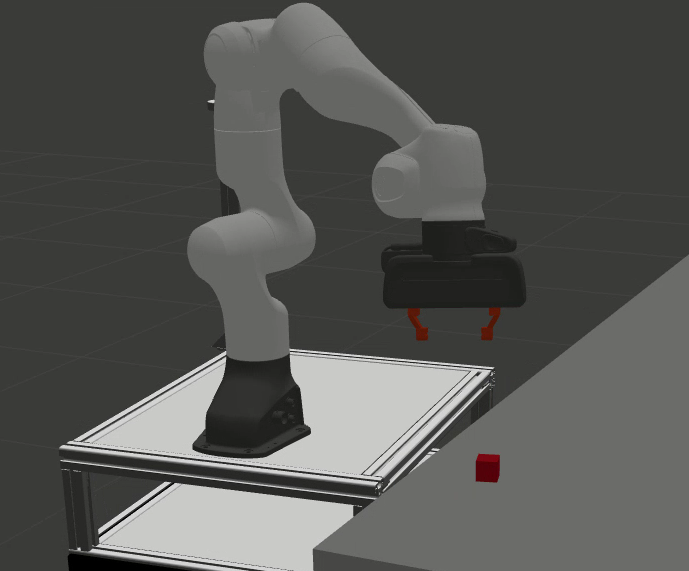} \hspace{0.1cm}
\includegraphics[height = 0.2\linewidth]{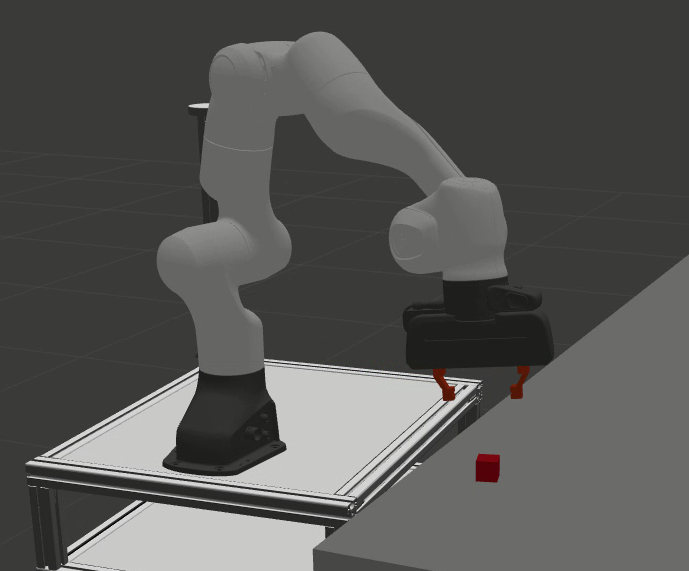} \hspace{0.1cm}
\includegraphics[height = 0.2\linewidth]{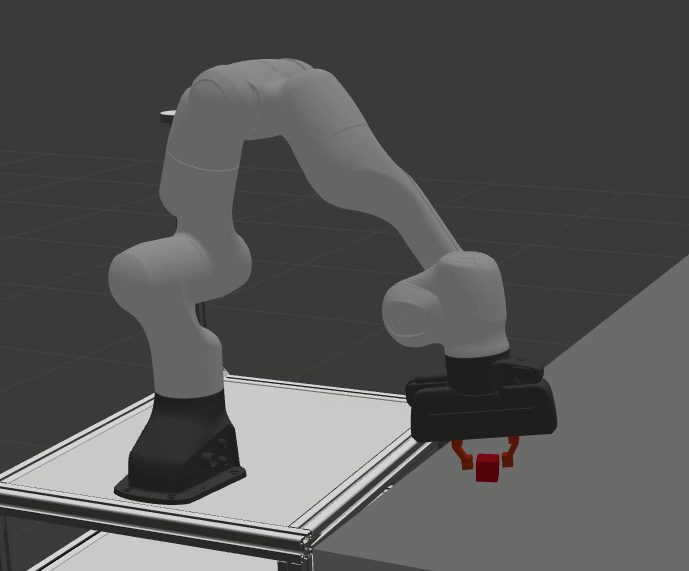} \hspace{0.1cm}
\label{fig:task-execution-picking} 
	}
\vspace{-0.25cm}

\subfigure[]{
\includegraphics[height = 0.2\linewidth]{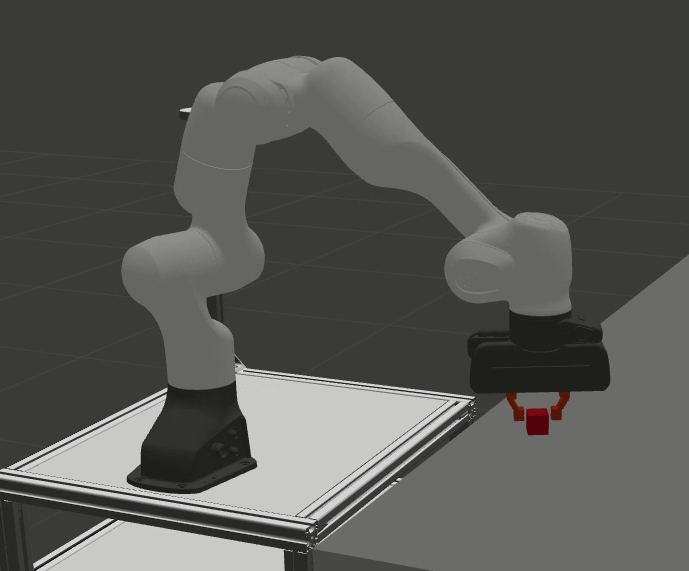} \hspace{0.1cm}
\includegraphics[height = 0.2\linewidth]{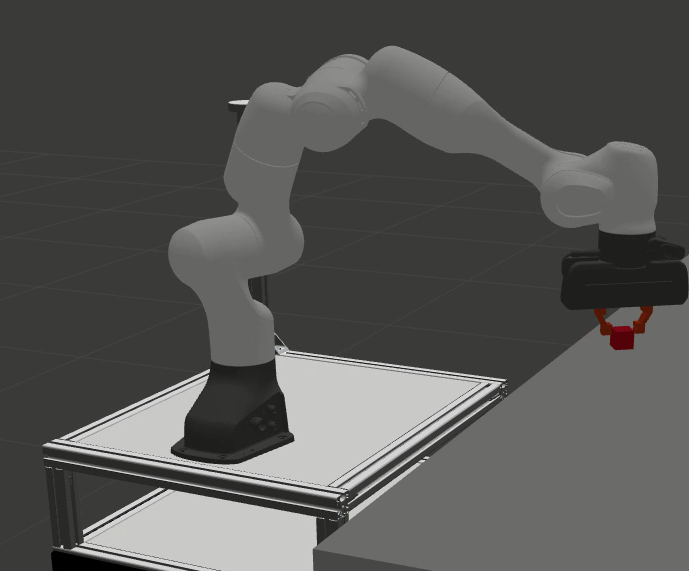} \hspace{0.1cm}
\includegraphics[height = 0.2\linewidth]{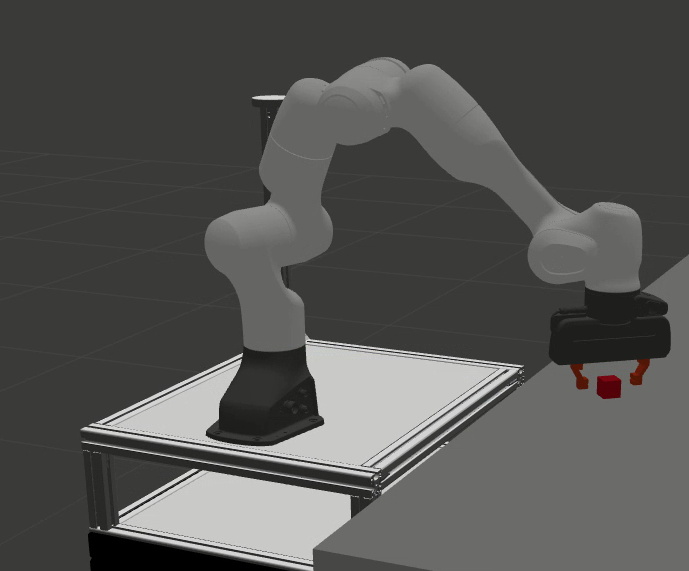} \hspace{0.1cm}
\label{fig:task-execution-placing} 
}
\vspace{-0.25cm}

\caption{Execution in the Gazebo simulator \cite{koenig2004design} of the pick and place operation described in Fig. \ref{fig:task-description-bt-2} and \ref{fig:task-description-bt-3}, with a Franka Emika Panda 7-DOF manipulator installed on a 4-wheel SUMMIT-XL STEEL mobile platform. \ref{fig:task-execution-picking} shows the sequence of the picking operation, while \ref{fig:task-execution-placing} the placing one.}
\label{fig:task-execution}
\vspace{-0.25cm}
\end{figure*}

We test our approach in simulation with a Franka Emika Panda 7-DOF manipulator installed on a 4-wheel SUMMIT-XL STEEL mobile platform. We build upon the implementation of the combined SoT and BT framework used in \cite{btsot}.
We design a task that consists of approaching a table, picking up and placing a 30mm cube from one side of the table to the other, while the mobile platform moves between the extremes of the table.
Fig. \ref{fig:task-description-bt} provides an overview of the task execution and its relative BT. 
For brevity, we do not show the entire subtree for the pick and place operation, which we designed according to the methodology described in \cite{btsot}.
The robot starts in position \textit{A}.
The Action Node \textit{Move to C} is executed to make the robot approach the table, while the Condition Node \textit{Robot Close to the Table} prevents the \textit{Pick and Place} subtree from being executed (Fig. \ref{fig:task-description-bt-1}).
Once the robot reaches the position \textit{B}, the condition is met and the \textit{Pick and Place} subtree is ticked while the Action \textit{Move to C} is still running (Fig. \ref{fig:task-description-bt-2}).
When the robot reaches the end of the table (position \textit{C}), the Action \textit{Move to B} is executed without affecting the right part of the tree (Fig. \ref{fig:task-description-bt-3}).
The Decorator Node \textit{Repeat} makes the platform keep on moving between the positions \textit{B} and \textit{C} until the pick and place operation is completed.

For the control of the mobile platform we use a standard differentiation kinematic controller and we keep the velocity constant in module, switching only its direction to move the platform between the positions \textit{B} and \textit{C}.
The modular nature of BTs has allowed us to design the BT and tune the controllers parameters in isolation for the two operations performed by the robotic arm (i.e., picking and placing) and the one performed by the mobile platform (i.e., movement between the two positions).
However, a further tuning of the SoT parameters was needed to make the pick and place operation work when combined with the platform movement.
In Fig. \ref{fig:task-execution} we show the execution in the Gazebo simulator \cite{koenig2004design} of the pick and place operation described in Fig. \ref{fig:task-description-bt-2} and \ref{fig:task-description-bt-3}.

The task taken into account allows to test the framework in a context that requires a basic form of coordination between the two involved controllers.
On the one hand, thanks to the BT, we can easily manipulate the coordination under the same control policy.
On the other hand, the use of the framework proposed in \cite{btsot} allows to inherit its advantages in terms of modularity and transparency.
Moreover, the platform movement during the pick and place simulates a possible navigation task that is typically performed by a mobile manipulator and makes the pick and place operation more challenging for the robot.

\section{\uppercase{Challenges}}
\label{sec:challenges}

The task described in Sec. \ref{sec:approach-evaluation} for evaluating the proposed approach highlights two open challenges which are essentially related to the parameters tuning and the policy design.

As previously mentioned, we needed to additionally refine the parameters of the SoT when performed in parallel with the platform movement.
This aspect underlines that, while the BT nature improves the policy modularity from the design point of view, the combination of different controllers brings inevitably some dependence in the parameters tuning.
Thus, the re-usability of subtrees in the BT is limited by the disadvantage of having to tune anyway the parameters involved, which becomes even harder when combining multiple controllers, given to the combinatorial explosion.

Moreover, in general the policy design is not as trivial as for the simple task considered in this paper.
Designing the BT to coordinate multiple controllers to achieve multiple goals at the same time is extremely complex, especially when dealing with dynamic environments, as it becomes difficult to predict all possible scenarios in advance.
Recent works have showed promising results that apply different methodologies to learn both the parameters and the structure of a BT \cite{9636292, learning-colledanchise-2019, 8794104}.
Our framework opens new possibilities for applying learning methodologies to address the above-mentioned challenges.

%%%%%%%%%%%%%%%%%%%%%%%%%%%%%%%%%%%%%%%%%%%%%%%%%%%%%%%%%%%%%%%%%%%%%%%%%%%%%%%%

%%%%%%%%%%%%%%%%%%%%%%%%%%%%%%%%%%%%%%%%%%%%%%%%%%%%%%%%%%%%%%%%%%%%%%%%%%%%%%%%
\balance
%%%%%%%%%%%%%%%%%%%%%%%%%%%%%%%%%%%%%%%%%%%%%%%%%%%%%%%%%%%%%%%%%%%%%%%%%%%%%%%%
%\newpage
\bibliographystyle{template/IEEEtran}
\bibliography{template/IEEEabrv,references}
\end{document}